\documentstyle{article} 
\begin{document}
\title{The {\it \lq \lq Fodor''}-FODOR fallacy bites back}
\author{Yorick Wilks}
\date{}
\maketitle
 {\parindent 0pt
\begin{abstract} 

The paper argues that Fodor and Lepore are misguided in their
attack on Pustejovsky's {\it Generative Lexicon}, largely because their
argument rests on a traditional, but implausible and discredited, view of the
lexicon on which it is effectively empty of content, a view that stands in the
long line of explaining word meaning (a) by ostension and then (b) explaining
it by means of a vacuous symbol in a lexicon, often the word itself after
typographic transmogrification.  (a) and (b) both share the wrong belief that
to a word must correspond a simple entity that is its meaning.  I then turn to
the semantic rules that Pustejovsky uses and argue first that, although they
have novel features, they are in a well-established Artificial Intelligence
tradition of explaining meaning by reference to structures 
that mention other structures assigned to
words that may occur in close proximity to the first.  It is argued that
Fodor and Lepore's view that there cannot be such rules is without foundation,
and indeed systems using such rules have proved their practical worth in
computational systems.  Their justification descends from line of argument,
whose high points were probably Wittgenstein and Quine that meaning is not to
be understood by simple links to the world, ostensive or otherwise, but by the
relationship of whole cultural representational structures to each other and
to the world as a whole\footnote{``One might say:  the ostensive definition
explains the use --the meaning-- of a word when the overall role of the word
in language is clear.  Thus if I know that someone means to explain a
colour-word to me the ostensive definition ``That is called {\it `sepia'}''
will help me to understand the word'' (Wittgensein, 1953, p.  30).}.
\end{abstract}
}

\section{Introduction}

Fodor and Lepore (FL from here on) have saddled up recently and ridden again
at the Windmills of Artificial Intelligence (AI):  this time against
Pustejovsky's {\it Generative Lexicon} (Pustejovsky, 1995:  FL call the work
TGL), so as to make an example for the rest of us.  I want to join in because
FL claim he is part of a wider movement they call {\it Informational Role
Semantics} (which I will call IRS as they do), and I count myself a long term
member of that movement.  But their weapons are rusty:  they wave about as
their sword of truth an old and much satirised fallacy, which Ryle (1957)
called the {\it ``Fido''}-Fido fallacy:  that to every word corresponds a
meaning, be it abstract denotation (as for FL), a concept, or a real world
object.  The special quality of the fallacy is the simple one-to-one mapping,
and not the nature of what corresponds to a word.

In the first part of this paper I want to show that the fallacy cannot be
pressed back into service:  it is old and overexposed.  It is important to do
this (again) even though, as the paper progresses, FL relent a little about
the need for the fallacy, and even seem to accept a part of the IRS position.
But they do this as men convinced that, really and truly and after all
concessions are made, the fallacy is still true.  It is not, and this needs
saying yet again.  In the second part of the paper, I will briefly touch on
issues specific to Pustejovsky's (JP) claims; only briefly because he is
quite capable of defending his own views.  In the third and final part I will
make some points to do with the general nature of the IRS position, within AI
and computational natural language processing, and argue that the concession
FL offer is unneeded:  IRS is a perfectly reasonable doctrine in its own right
and needs no defence from those who really believe in the original fallacy.

\section{{\it ``Fido''} and FIDO}

Fodor and Le Pore's dissection of JP's book is, and is intended to be, an
attack on a whole AI approach to natural language processing based on symbolic
representations, so it is open to any other member of that school to join in
the defence.  IRS has its faults but also some technological successes to show
in the areas of machine translation and information extraction (e.g.  Wilks et
al., 1993), but is it well-founded and philosophically defensible?

Many within IRS would say that does not matter, in that the only defence
lexical or other machine codings need in any information processing system is
that the system containing them works to an acceptable degree; but I would
agree with those who say it is defensible, or is at least as well founded as
the philosophical foundation on which FL stand.  That is, I believe, one of
the shakiest and most satirised of this century, and loosely related to what
what Ryle (1957) called the {\it ``Fido''}-Fido fallacy:  the claim that to
every word corresponds a concept and/or a denotation, a view that has crept
into everyday philosophical chat as the joke that the meaning of life is life'
(life prime, the object denoted by {\it ``life''}\footnote{`Fido' or Fido-prime
are common notations for denotations corresponding to words.  FL seem to
prefer small caps FIDO, and I will use that form from their paper in the
argument that follows.}.

It is a foundation of the utmost triviality, that comes from FL
(op.cit., p.1) in the form:\\

(1) The meaning of {\it ``dog''} is DOG.\\

{\parindent 0pt

They seem to embrace it wholeheartedly, and prefer it to any theory,
like TGL, offering complex structured dictionary entries, or even any
paper dictionary off a shelf, like Webster's, that offers even more
complex structures than TGL in a form of English.  FL embrace an empty
lexicon, willingly and with open eyes: one that lists just DOG as the
entry for {\it ``dog''}.  The questions we must ask, though the answer is
obviously {\it no} in each case, are:
}

\begin{itemize}

\item is (1) even a correct form of what FL wants to say?

\item could such a dictionary consisting of statements like (1) serve
any purpose whatever, for humans or for machines?

\item would one even need to write such a dictionary, supposing one
believed in a role for such a compilation, as opposed to say, 
saving space by storing one as a
simple rule for capitalizing any word whose meaning was wanted?
\end{itemize}

{\parindent 0pt

The first of these points brings back an age of linguistic analysis
contemporary with Ryle's, in particular the work of writers like Lewy
(1964); put briefly, the issue is whether or not (1) expresses anything
determinate (and remember it is the mantra of the whole FL paper), or
preferable to alternatives such as:\\
}

(2) The meaning of {\it ``dog''} is a domestic canine animal.\\

{\parindent 0pt

or\\
}

(3) The meaning of {\it ``dog''} is a dog.\\

{\parindent 0pt

or even\\
}

(4) The meaning of {\it ``dog''} is {\it ``domestic canine animal''}.\\

{\parindent 0pt

not to mention\\
}

(5) The meaning of {\it ``dog''} is {\it ``dog''}.\\

{\parindent 0pt

The two sentences (2) and (3) are perfectly sensible, depending on the
circumstances: (2) is roughly what real, non-Fodorian, dictionaries
tell you, which seems unnecessary for dogs, but would be more
plausible if the sentence was about marmosets or wombats.  (3) is
unhelpful, as it stands, but perhaps that is accidental, for if we
translate it into German we get something like :\\
}

(3a) Die Bedeutung von {\it ``dog''} ist ein Hund.\\

{\parindent 0pt

which could be very helpful to a German with little or no knowledge of
English, as would be\\
}

(2a) Die Bedeutung von {\it ``dog''} ist ein huendliche Haustier.\\  

{\parindent 0pt

To continue with this line of argument one needs all parties to accept the
reality of translation and its role in argument:  that there are translations,
at least between close languages for simple sentences, and no amount of
philosophical argument can shift that fact.  For anyone who cannot accept
this, there is probably no point in arguing about the role of lexicons at all.
}

Both (2) and (3), then, are sensible and, in the right circumstances,
informative:  they can be synonymous in some functional sense since both, when
translated, could be equally informative to a normal fluent speaker of another
language.  But (4) and (5) are a little harder:  their translations would be
uninformative to a German when translated, since translation does not
translate quotations and so we get forms like:\\

(5a) Die Bedeutung von {\it ``dog''} ist {\it ``dog''}.\\

{\parindent 0pt

and similarly for a German (4a) version of the English (4).  
These sentences therefore cannot be synonymous with
(3) and (2) respectively.  But (4) might be thought no more than an odd form
of a lexical entry sentence like (3), spoken by an English speaker.
}

But what of (1); who could that inform about anything?  Suppose we
sharpen the issue by again asking who its translation could inform and
about what:\\

(1a) Die Bedeutung von {\it ``dog''} ist DOG.\\

{\parindent 0pt

(1a) tells the German speaker nothing, at which point we may be told that DOG
stands for a denotation and its name is arbitrary.  But that is just old
nonsense on horseback:  it implies that the English speaker cannot understand
(1) either, since DOG might properly be replaced by G00971 if the final symbol
in (1) is truly arbitrary.  It is surely (3), not (1), that tells us what the
denotation of {\it ``dog''} is, in the way language is normally used to do
such things.
}

DOG in (1) is simply a confidence trick:  it is put to us as having the role
of the last term in (3).  When and only when it is in the same language as the
final symbol of (3) (a fact we are confidently assured is arbitrary) it does
appear to point to dogs.  However, taken as the last term in the synonymous
(1a) it cannot possibly be doing that for it is incomprehensible, and
functioning as an (untranslated) English word, exactly as in the last term of
(5).  But, as we saw, (5) and (3) cannot be synonymous, and so DOG in (1) has
two incompatible roles at once, which is the trick that gives (1)
interpretations that flip uncontrollably between the (non-synonymous) (3) and
(5).  It is an optical illusion of a sentence.

In conclusion, then, (1) is a dangerous sentence, one whose upper
case-inflation suggests it has a function but which, on careful examination,
proves not to be there:  it is either (case-deflated) a form of the
commonsense (3), in which case it loses its capitals and all the role FL
assign to it, since it is vacuous in English, or just a simple bilingual
dictionary entry in German or some other language.  Or it is a form or (5),
uninformative in any language or lexicon but plainly a triviality, shorn of
any philosophical import.

Those who still have worries about this issue, and wonder if capitalizing may
not still have some merit, should ask themselves the following question:
which dog is the real DOG?  The word {\it ``dog''} has 24 entries even in a
basic English dictionary like Collins, so how do FL know which one is intended
by (1)?  If one is tempted to reply, well DOG will have to be subscripted
then, as in DOG$_{1}$, DOG$_{2}$ etc, then I shall reply that we will then be
into another theory of meaning, and not one of simple denotations.  My own
suspicion is that all this can only be understood in terms of Fodor's {\it
Language of Thought} (1975) and that DOG for FL is a simple primitive in that
language, rather than a denotation in the world or logical space.  However, we
have no access whatever to such a language, though Kay among others has given
arguments that, if anything like an LOT exists, it will have to be subscripted
(Kay, 1989), in which case the role of (1) will have to be rethought from
scratch.  All the discussion above will still remain relevant to such a
development, and the issue of translation into LOT will then be the key one.
However, until we can do that, and in the presence of a LOT native speaker, we
may leave that situation aside and await developments.

The moral for the rest of the discussion, and the role of IRS and TGL, is
simple:  some of the sentences numbered above are like real, useful, lexical
entries:  (3) is a paradigm of an entry in a bilingual lexicon, where
explanations are not crucial, while (2) is very like a monolingual lexical
entry, where explanations are the stuff of giving meaning.

\section{Issues concerning TGL}

The standard of the examples used by FL to a attack TGL is not  at all
challenging; they claim that JP's:\\ 

(6) He baked a cake.\\

{\parindent 0pt

is in fact ambiguous between JP's {\it create} and {\it warm up} aspects of
{\it ``bake''}, where baking a cake yields the first, but baking a potato the
second.  JP does not want to claim this is a sense ambiguity, but a systematic
difference in interpretation given by inferences cued by features of the two
objects, which could be labels such as ARTIFACT in the case of the cake but
not the potato.
}

``But in fact, {\it bake a cake} is ambiguous.  To be sure, you can make a
cake by baking it; but also you can do to a (pre-existent) cake just
what you can do to a (pre-existent) potato: viz. put it in the oven
and (non creatively) bake it.''  (op.cit. p.7) 

From this, FL conclude, {\it ``bake''} must be ambiguous, since {\it ``cake''}
is not.  But all this is absurd and untrue to the simplest facts.  Of course,
warming up a (pre-existent) cake is not baking it; whoever could think it
was?  That activity would be referred to as warming a cake up, or through,
never as baking it.  You can no more bake a cake again, with the other
interpretation, than you can bake a potato again and turn it into an artifact.
FL like syntactically correlated evidence in semantics, and they should have
noticed that a {\it baked potato} is fine but a {\it baked cake} is not, which
correlates with just the difference JP requires (cf.  baked fish/meat).

It gets worse:  FL go on to argue that if ARTIFACTs are the distinguishing
feature for JP then {\it bake a trolley car} should take the creative reading
since it is an artifact, completely ignoring the fact that the whole JP
analysis is based on the (natural) assumption that potatoes and cakes both
share some attribute like FOOD (as trolley cars do not) which is the only way
the discussion can get under way:  being a FOOD is a necessary condition for
this analysis of {\it ``bake''} to get under way.

FL's key argument against TGL is that it is not possible to have a rule, of
the sort JP advocates, that expands the content or meaning of a word in virtue
of (the meaning content of) a neighbouring word in a context, namely, a word
in some functional relation to the first.  So, JP, like many in the IRS
tradition, argues that in:\\

(7) John wants a beer.\\

{\parindent 0pt

the meaning of {\it ``wants''} in that context, which need not be taken to be
any new or special or even existing sense of the word, is to be glossed as
{\it wants to drink} a beer.  This is done by a process that varies in detail
from IRS researcher to researcher, but comes down to some form of:\\
}

(8) X wants Y $\Rightarrow$ X wants to do with Y whatever is normally done
with Y.\\

{\parindent 0pt

An issue over which AI researchers have differed is whether this knowledge of
normal or default use is stored in a lexical entry or in some other
computational knowledge form such as one that was sometimes called a script
(Schank and Abelson, 1997) and thought of as indexed by words but was much
more than a lexical entry.  It is not clear that one needs to discriminate
between structures, however complex, if they are indexed by a word or words.
JP stores the inference captured in (8) within the lexical entry under a label
TELIC that shows purpose.  In earlier AI systems, such information about
function might be stored as part of a lexical semantic formulas attached to a
primitive GOAL (Charniak and Wilks, 1976\footnote{ 
In preference semantics (Wilks) {\it ``door''} was coded as a formula 
(that could be displayed as a binary tree) such as:\\
((THIS((PLANT STUFF)SOUR)) ((((((THRU PART)OBJE) (NOTUSE *ANI))GOAL) ((MAN
USE) (OBJE THING))))\\ where the subformula:\\
((((THRU PART)OBJE) (NOTUSE *ANI))GOAL)\\
was intended to convey that the function of a door is to prevent passage
through an aperture by a human.}), or (depending on its salience) within an
associated knowledge structure\footnote{Such larger knowledge structures were
called pseudo-texts (Wilks) in preference semantics (to emphasize the
continuity of language and world knowledge):  one for {\it ``car''} (written
[car]) would contain a clause like [car consume gasoline] where each lexical
item in the pseudo-text was an index to a semantic formula (in the sense
of note 3) that explicated it.}.  Some made the explicit assumption that
a system should be sufficiently robust that it would not matter if such
functional information was stored in more than one place in a system, perhaps
even in different formats.
}

For FL all this is unthinkable, and they produce a tortuous argument
roughly as follows:

\begin{itemize}

\item {\it ``Fido''}-FIDO may not be the only form for a lexicon, but an
extension could only be one where an expansion of meaning for a term
was independent of the control of all other terms, as it is plainly
not in the case of JP's (8)).

\item Any such extension would be to an underlying logical form, one
that should also be syntactically motivated.
\end{itemize}

{\parindent 0pt

FL then produce a complex algorithm (op.cit.  p.10) that expands {\it
``want''} consistently with these assumptions, one which is hard to follow and
comes down to no more than the universal applicability (i.e.  if accepted it
must be applied to all occurrences of {\it ``want''} regardless of lexical
context) of the rule:\\
}

(9)  X wants Y $\Rightarrow$ X wants to have Y.\\

{\parindent 0pt

This, of course, avoids, as it is intended to, any poking about in the
lexical content of Y.  But it is also an absurd rule, no matter what
dubious chat is appended to it about the nature of ``logical form''.
Consider:\\
}

(10a)	I want an orange.\\

(10b)   I want a beer.\\

(10c)	I want a rest.\\

(10d)	I want love.\\

{\parindent 0pt

(10a) and (10b) seem intuitively to fit the IRS rule (8) and the FL
rule (9).  (10c) might conform to some appropriate IRS coding to
produce (from (8)): X wants to experience a rest, but the apparently
felicitous application of FL's (9), yielding X wants to have a rest,
is purely fortuitous, since {\it have a rest} is a lexicalised form having
nothing at all to do with possession, which is the only natural
interpretation of (9).  This just serves to show the absurdity of FL's
``content-free'' rule (9) since its application to (10c) cannot possibly
be interpreted in the same way as it was when producing {\it X wants to
have a beer}.
}

Only the IRS rule (8) distinguishes appropriate from inappropriate
applications of rules to (10c).  One could make the same case with
(10d), where the FL rule (9) produces only ambiguous absurdity, and
the applicability of the IRS rule (8) depends entirely on how the
function of {\it ``love''} has been coded, if at all.  However, the purpose of
this section has been  not to defend an IRS view or rule (8) in
particular, but to argue that there is no future at all in FL's
grudging, context free, rule (9), in part because it is context free. 

JP's specific claim is not that the use of rule (8) produces a new
sense, or we would have a new sense corresponding to many or most of
the possible objects of wanting, a more promiscuous expansion of the
lexicon (were it augmented by every rule application) than would be
the case for {\it bake a potato/cake} where JP resisted augmentation of
the lexicon, though other researchers would probably accept it.  Nor
is this like the application of ``normal function'' to the
transformation of\\ 

(11) My car drinks gasoline.\\

\noindent
in (Wilks, 1980) where it was suggested that {\it ``drink''} should be
replaced by the structure for {\it ``consume''} (as in note 4 below) in a
context containing broken preferences (unlike (7)) and where augmentation of
the lexicon would be appropriate (like (11)) and if the ``relaxation'', as some
would call it, became statistically significant, as it has in the case of
(11).

It is not easy to pin down quite why FL find the rule (8) so objectionable,
since their rule (9), like (8), is not, as they seem to believe, distinguished
by logical form considerations.  The traditional (Quine, 1953) logical opacity
of {\it ``want''} is such that inferences like (8) and (9) can never be
supported by strong claims about logical form whose transformations must be
deductive, and one can always want X without wanting Y, no matter what the
logical relations of X and Y.  Hence, neither (8) nor (9) are deductive rules,
and FL have no ground in context-dependence to swallow the one but not the
other.

Contrary to what FL seem to assume, an NLP algorithm to incorporate or
graft part of the lexical entry for one word (e.g. {\it ``beer''}) into
another (e.g. {\it ``want''}) is not practically difficult.  The only issue
for NLP research is whether and when such inferences should be drawn:
at first encounter with such a collocation, or when needed in later
inference, a distinction corresponding roughly to what is
distinguished by the oppositions forward and backward inference, or
data-driven and demand-driven inference.  This issue is connected with
whether a lexical entry should be adapted rather than a database of world
knowledge and, again contrary to FL's assumptions, no NLP researcher
can accept a firm distinction between these, nor is there one, any
more than a firm analytic-synthetic distinction has survived decades
of scepticism.

One can always force such a distinction into one's system at trivial cost, 
as Carnap (1947) did with his formal and material modes of
sentences containing the same information:\\

(12f) {\it ``Edelweiss''} has the property {\it ``flower''}.\\

(12m) An Edelweiss is a flower.\\

{\parindent 0pt
but the distinction is wholly arbitrary\footnote{Provided one remembers 
always that forms like:\\
{\it ``Edelweiss''} has nine letters\\
is in material mode even though it could look like the formal mode.
The formal mode of what it expresses is:\\
{\it ``Edelweiss''} has the property {\it ``nine letters''}.}.\\

JP's treatment of more structural intensional verbs like {\it ``believe''} is
far more ingenious than FL would have us believe, and an interesting
advance on previous work: it is based on a richer notion of default
than earlier IRS treatments.  JP's position, as I understand it, is
that the default, or underlying, structure associated with {\it ``believe''}
is:\\
}

X believes p\\

{\parindent 0pt

where p is expanded by default by the rule:\\
}

(13) X believes p $\Rightarrow$ X believes (Y communicates p).\\

{\parindent 0pt

FL of course object again to another expansion beyond their
self-imposed limit of context-freeness for which, as we saw, there is
no principled defence, while failing to notice that (13) is in fact
context-free in their sense.
}

For me, the originality of (13) is not only that it can expand forms
like:\\

(14) John believes Mary.\\

{\parindent 0pt

but can also be a general (context-free) default, overriding forms
like:\\
}

(15) John believes pigs eat carrots.\\

{\parindent 0pt

in favour of the more complex:\\
}

(16) John believes (Y communicates (pigs eat carrots)).\\

{\parindent 0pt

which is an original expansion according to which all beliefs can be seen as
the result of some communication, often from oneself (when Y = John in (16)).
There certainly were default expansions of {\it ``believe''} in IRS before JP
but not of this boldness\footnote{In preference semantics (Wilks, 1972) {\it
``believe''} had a lexical entry showing a preference for a propositional
object, so that ``John believes Mary'' was accepted as a preference-breaking
form but with a stored expansion of the object in the lexical entry for {\it
``believe''} of a simple propositional form (Mary DO Y) with what is really 
an empty verb variable DO, and not a communication specifically act like TGL.}.
}

\section{Some general IRS principles}

Once (Wilks, 1971, 1972) I tried to lay out principles for something very like
IRS, and which still seem to underlie the position arrived at in this
discussion; it would be helpful for FL to see IRS not simply as some form of
undisciplined, opportunistic, discipline neighbouring their own professional
interests.  Let me restate two of these principles that bear on this
discussion:

\begin{itemize}

\item Meaning, in the end, is best thought of as other words, and that is the
only position that can underpin a lexicon- and procedure-based approach to
meaning, and one should accept this, whatever its drawbacks, because the
alternative is untenable and not decisively supported by claims about
ostension.  Quine (1953) has argued a much more sophisticated version of this
for many years, one in which representational structures are only compared
against the world as wholes, and local comparisons are wholly symbolic.
Meanings depend crucially upon explanations and these, formally or
discursively, are what dictionaries offer.  This solution to the problem is
indeed circular, but not viciously so, since dictionaries rarely offer small
dictionary circles (Sparck Jones, 1966) like the classic, and unsatisfying,
case where {\it ``furze''} is defined as {\it ``gorse''} and vice versa.
Meanings, in terms of other words, is thus a function of circle size:  furze
gorse is pathological, in the sense of unhelpful, yet, since a dictionary
definition set is closed, and must be circular, not all such circles can be
unhelpful or dictionaries are all and always vacuous.

On the other hand, FL's original position of the section 2 above, is not
really renounced by the end of their paper, and is utterly untenable,
not only for the analytic reasons we have given, but because it could
not form the basis of any possible dictionary, for humans (seeking
meaning explanations) or for NLP.

Indeed, as we pointed out earlier, no lexicon is needed for the
{\it ``dog''}-DOG theory, since a simple macro to produce upper-case forms
will do, give or take a little morphological tweaking for the
{\it ``boil''}-BOILING cases.

\item Semantic well-formedness is not a property that can decidably be
assigned to utterances, in the way that truth can to formulas in parts
of the predicate calculus, and as it was hoped for many years that
``syntactically-well formed'' would be assignable to sentences.

This point was made in some detail in (Wilks, 1971) on the basis that no
underlying intuition is available to support semantic
well-formedness\footnote{This property must intuitively underlie all
decidability claims and procedures:  Goedel's proof that there are true but
undecidable sentences in a class of calculi only makes sense on the assumption
that we have some (intuitive) way of seeing those sentences are true (outside
of proof)}, since our intuitions are dependent on the state of our (or our
machine's) lexicons when considering an utterance's status, and that we are
capable of expanding our lexicons (in something like the ways discussed in
this paper) so as to bring utterances iteratively within the boundary of
semantic well-formedness, and in a way that has no analogy in truth or syntax.
Thus, no boundary drawing, of the sort required for a decidable property, can
be done for the predicate Semantically-well-formed.  Belief in the opposite
seems one of the very few places where JP and FL agree, so further discussion
may prove necessary.
\end{itemize}

{}

\end{document}